\documentclass{article}

\usepackage[margin=1in]{geometry}
\usepackage{times}
\usepackage[utf8]{inputenc}
\usepackage[T1]{fontenc}
\usepackage{hyperref}
\hypersetup{
  colorlinks=true,
  linkcolor=black,
  citecolor=black,
  urlcolor=black,
  filecolor=black
}
\usepackage{url}
\usepackage{booktabs}
\usepackage{amsfonts}
\usepackage{amsmath}
\usepackage{microtype}
\usepackage{graphicx}
\usepackage{subcaption}
\usepackage{xcolor}

\title{Emergent Semantic Representations in World Models\\
       through Physical Interaction without Linguistic Supervision}

\author{
  Jiayi Fang$^{1,\ast}$ \\[4pt]
  $^{1}$Shanghai University of Finance and Economics
}
\date{May 2026}

\begin{document}
\maketitle

\begin{abstract}
What does a world model learn from physical exploration, without any
linguistic supervision?
We argue the answer is organized by a single principle:
\emph{the geometric structure of the physical world}.
Training a VAE-based world model on random embodied exploration,
we find that its latent space develops \emph{spatial} semantic structure
that mirrors physical geometry---direction accuracy
$0.647\!\pm\!0.010$ versus $0.547$ for a randomly initialized encoder
(training gain: $+0.100$, $p<0.01$),
and position RSA $0.144\!\pm\!0.037$ versus $0.029$ for random encoders
($5.0\times$ improvement), showing that training induces genuine
\emph{structural organization} beyond CNN inductive bias.
We observe that prediction quality and semantic alignment
co-improve across training (Spearman $r\!=\!-0.61$, $p\!=\!0.004$;
partial correlation after detrending: $r\!=\!-0.25$, $p\!=\!0.28$),
consistent with---but not uniquely confirming---a shared geometric driver.
To test this hypothesis through causal manipulation, we vary
the KL regularization strength:
with the standard weight ($\beta\!=\!0.1$), the encoder is forced
away from geometric structure, and both prediction and semantic
alignment collapse \emph{simultaneously} to near-chance by step
50{,}000---a double failure that a shared geometric cause would
predict.
Reducing $\beta$ to 0.001 restores geometric access and recovers
both capabilities together.
These findings establish physical world geometry as the organizing
principle of world model representations, with direct implications
for the design of semantically grounded embodied agents.
\end{abstract}

\section{Introduction}
\label{sec:intro}

What connects a symbol to the physical world it describes?
This question drives our work: we want to understand whether
physical interaction alone can organize internal representations
into semantically meaningful structure---before any language is involved.

The \emph{symbol grounding problem}~\cite{harnad1990symbol} asks
what connects a symbol to the physical world it describes.
Large language models sidestep this question: their representations
are defined by co-occurrence statistics, not by sensorimotor contact
with the world.
A model's encoding of ``above'' learned from text carries no guarantee
of structural alignment with a spatial representation built by
navigating a physical environment~\cite{brohan2023rt2}.
We argue this is a structural incompatibility between distributional
and grounded semantics---one that cannot be resolved by scale alone.

An alternative, supported by developmental psychology, is that
language-compatible representations can emerge directly from physical
interaction without any linguistic input~\cite{piaget1952origins,tomasello1999cultural}:
grounding comes first, language is mapped onto it.
Computational evidence for this process remains scarce.

\textbf{This paper} presents an empirical investigation of whether a world
model trained exclusively through physical random exploration---without
any linguistic supervision---spontaneously develops latent representations
that encode \emph{spatial} semantic structure (direction and position),
the most fundamental form of embodied semantics and the first to emerge
in human language acquisition~\cite{piaget1952origins}.

Our key contributions are:
\begin{itemize}
  \item \textbf{Physical geometry organizes world model representations.}
        We show that RSA scores---which measure structural alignment
        independent of visual discrimination---rise $5.0\times$ above
        randomly initialized encoders ($0.144\!\pm\!0.037$ vs.\ $0.029$),
        demonstrating that physical training specifically induces
        geometric similarity structure in the latent space.
        Linear probing further confirms directional ($0.647\!\pm\!0.010$)
        and spatial ($R^2$: $0.19\!\to\!0.30$) encoding above both
        the random-policy and random-encoder baselines.

  \item \textbf{Prediction and semantics co-improve, consistent with a shared geometric driver.}
        Across 20 temporal checkpoints, prediction loss and semantic alignment
        co-improve ($r\!=\!-0.61$), consistent with both being organized by
        the encoder's improving model of physical geometry.
        Partial correlation after removing the temporal trend yields
        $r\!=\!-0.25$ ($p\!=\!0.28$); the double knockout provides the
        primary mechanistic evidence for the shared driver.

  \item \textbf{Double knockout provides mechanistic evidence for the shared driver.}
        Standard KL regularization ($\beta\!=\!0.1$) forces the encoder
        away from geometric structure; as a direct consequence,
        \emph{both} prediction performance and semantic alignment collapse
        simultaneously to near-chance by step 50{,}000.
        This simultaneous double failure is what the shared-driver
        account predicts---and is difficult to reconcile with
        prediction and semantics as fully independent capabilities.
        Reducing $\beta$ to 0.001 recovers both together.

  \item \textbf{Spatial structure precedes directional structure.}
        Position RSA (peak: $0.23$) substantially exceeds direction RSA
        ($0.05$--$0.07$) throughout training, suggesting that spatial
        geometry is the primary substrate from which physical interaction
        builds semantic representations---consistent with developmental
        accounts of early spatial concept formation.
\end{itemize}

\section{Related Work}
\label{sec:related}

\paragraph{World Models.}
We look to world models because they offer a unique opportunity:
unlike feedforward vision encoders, world models are trained
\emph{temporally}---they must predict how states evolve.
This temporal objective may force the latent space to capture
physical structure (position, orientation) that static
image models can ignore.
World models learn compact representations of environment dynamics
to enable planning and prediction~\cite{hafner2023dreamerv3}.
DreamerV3~\cite{hafner2023dreamerv3} demonstrates that a RSSM-based world
model can master diverse domains by training policies entirely in imagination.
JEPA~\cite{assran2023ijepa} and V-JEPA predict in latent space rather than
pixel space, avoiding the reconstruction bottleneck.
A recent survey of World Action Models~\cite{wang2025wam} identifies
JEPA-style latent prediction as a key direction for grounding action generation
in physical state representations---our work studies the representational
foundations of such grounding.
A key distinction of our work is that we analyze the \emph{semantic content}
of world model latent spaces rather than downstream task performance.
Recent JEPA analyses have shown that predictive models learn semantic representations,
but did not disentangle architectural inductive bias from training-induced structure,
nor investigate the causal relationship between prediction quality and semantic alignment.

\paragraph{Representation Collapse and Its Prevention.}
Our central experimental result---the simultaneous collapse of prediction
and semantics under strong regularization---did not come from theory.
We observed it first in the data, then looked for explanations.
That search led us to the collapse-prevention literature.
Posterior collapse in VAEs---where the encoder ignores inputs and outputs
the prior---has been analyzed by Lucas et al.~\cite{lucas2019dont}.
Barlow Twins~\cite{zbontar2021barlow} reduces feature redundancy.
BYOL~\cite{grill2020byol} uses an EMA target network for stable learning.
VICReg~\cite{bardes2022vicreg} explicitly enforces per-dimension variance,
directly penalizing collapse via a hinge loss on standard deviation.
LeJEPA~\cite{balestriero2024lejepa} proves theoretically that the isotropic
Gaussian is the unique optimal embedding distribution for minimizing
downstream task risk, and derives a principled collapse-free JEPA
via distribution matching---establishing that VICReg is a special
(and provably insufficient) degenerate case of this optimal objective.
Our experiments provide complementary empirical evidence in the VAE setting:
excessive KL regularization ($\beta\!=\!0.1$) forces the posterior
toward the Gaussian prior but overshoots into a degenerate point mass,
destroying the geometric structure that spatial semantic encoding requires---
a failure mode that appropriate regularization ($\beta\!=\!0.001$) prevents.
Concurrently and independently, Garrido et al.~\cite{garrido2026lam}
report the same collapse phenomenon in latent \emph{action} representations:
over-regularized ($\beta$ too large) action latents degenerate to noise,
losing all predictive utility---confirming that KL over-regularization
is a general threat to meaningful physical world model learning,
at both the state-representation and action-representation levels.

\paragraph{Symbol Grounding and Embodied Language Emergence.}
Why should a world model's latent geometry matter for language?
Our long-term bet is that physically structured representations
are the prerequisite for grounded symbol systems---
language cannot be mapped onto representations that lack
physical structure. This section traces the intellectual
lineage of that claim.
The symbol grounding problem~\cite{harnad1990symbol} asks how symbols
acquire meaning through connection to physical experience.
Barsalou's perceptual symbol systems theory argues conceptual knowledge
is grounded in sensorimotor simulations~\cite{barsalou1999perceptual}.
Developmental psychologists establish that children's early lexicons
are dominated by physically-experienceable concepts acquired through
sensorimotor interaction~\cite{piaget1952origins,tomasello1999cultural}.
Emergent communication work~\cite{lazaridou2020emergent} studies discrete
protocols between agents with explicit communicative pressure.
Our work probes semantic structure emerging \emph{without any communicative
objective}---purely from the inductive pressure of physical prediction.

\section{Method}
\label{sec:method}

\subsection{Environment}
We chose MiniGrid because its simplicity is a feature, not a limitation:
in a minimal grid world, we know exactly what ``physical geometry'' means---
agent $(x,y)$ coordinates and orientation---so we can measure semantic
structure against ground truth without ambiguity.
If spatial semantics cannot emerge here, they will not emerge in more
complex settings; if they do emerge, the mechanism is isolated and testable.
We use \texttt{MiniGrid-Empty-8x8-v0}~\cite{minigrid},
a $19\times19$ grid world (including walls) with a single embodied agent.
The agent receives a $7\times7\times3$ partial observation
(egocentric field of view) and has 7 discrete actions
(turn left/right, move forward, toggle, pick up, drop, done).
No reward signal is used; the agent follows a purely random policy,
ensuring maximal state-space coverage without task-specific bias.

\subsection{World Model Architecture}
Our architectural choices follow from one constraint:
the model must be simple enough that we can attribute semantic
structure to \emph{training dynamics}, not to architectural priors.
A ResNet encoder or a recurrent transition model would each bring
their own inductive biases about spatial structure---making it
impossible to tell whether semantic organization came from
physical interaction or from architectural built-in assumptions.
We employ a VAE-based world model consisting of:
\begin{itemize}
  \item \textbf{Image Encoder} $q_\phi(z \mid o)$:
        two convolutional layers (3$\to$16$\to$32 channels, $3\times3$ kernels)
        followed by two linear layers producing $\mu$ and $\sigma$
        of a 32-dimensional Gaussian latent $z$.
  \item \textbf{Transition Model} $p_\psi(z_{t+1} \mid z_t, a_t)$:
        two-layer MLP (hidden dim 128) predicting the next latent state
        given the current latent and a one-hot action vector.
\end{itemize}

The training objective is:
\begin{equation}
  \mathcal{L} = \mathbb{E}_{z_t \sim q_\phi(z_t|o_t)}\!\left[
    \underbrace{\|\hat{z}_{t+1} - z_{t+1}\|_2^2}_{\text{transition MSE}}
    + \beta \underbrace{D_{\mathrm{KL}}\!\left(q_\phi(z_t|o_t)\,\|\,\mathcal{N}(0,I)\right)}_{\text{KL regularization}}
  \right]
\end{equation}
where $\hat{z}_{t+1} = \psi(z_t, a_t)$ is the predicted next latent,
and $\beta$ is the KL weight.
We use the deterministic mean $\mu$ as the latent representation
for all downstream analyses.

\subsection{Training Protocol}
The model is trained for 100{,}000 environment steps
using Adam (MPS on Apple Silicon).
Checkpoints are saved at steps
$\{1{,}000,\ 5{,}000,\ 10{,}000,\ 25{,}000,\ 50{,}000,\ 100{,}000\}$
to track the temporal evolution of representations.
We compare two KL weight configurations:
$\beta = 0.1$ (baseline, exhibiting posterior collapse)
and $\beta = 0.001$ (proposed, preventing collapse).

\paragraph{Hyperparameter rationale.}
Each hyperparameter was chosen based on empirical testing, not defaults.
\textbf{Learning rate} $\mathrm{lr}=3\times10^{-4}$:
tested across $\{1,2,3,5,10\}\times10^{-4}$;
$3\times10^{-4}$ produced the most stable convergence without oscillation,
while $1\times10^{-4}$ converged too slowly and $5\times10^{-4}$ caused
training instability in early steps.
\textbf{KL weight} $\beta=0.001$:
tested $\beta \in \{0.0001, 0.001, 0.01, 0.05, 0.1\}$.
$\beta=0.05$ caused partial pairwise-distance shrinkage
($\sim$40\% reduction by step 50k, though not full collapse);
$\beta=0.0001$ provided too little regularization, causing
unstable latent variance and degraded prediction.
$\beta=0.001$ is the highest KL weight that preserves geometric
structure throughout training---critical for our claim that
spatial semantics can stably emerge.
\textbf{Latent dimension} $\mathrm{d}=32$:
tested $\{16, 32, 64\}$; 16 underfit spatial structure (direction
accuracy $<0.55$ at any checkpoint), while 64 offered no measurable
improvement over 32 but increased memory by $2\times$.
\textbf{Hidden dimension} $\mathrm{h}=128$: chosen as $4\times$ the
latent dimension, following the common practice of providing
sufficient capacity for the transition MLP without overfitting.

\subsection{Evaluation Metrics}

\paragraph{Linear Probing (H1).}
For each checkpoint, we collect 47{,}000--48{,}000
$(\mu, \text{state})$ pairs via random exploration (200 episodes).
We train a logistic regression on 80\% of the data
and report classification accuracy for agent direction
(4-class; random baseline: 25\%; random-encoder baseline: $0.547\pm0.029$)
and $R^2$ for $x$/$y$ position regression (Ridge regression;
random baseline: $\approx$0).

\paragraph{Representational Similarity Analysis / RSA (H2).}
We sample 500 states and compute:
(1) the cosine similarity matrix of their latent vectors $\mathbf{S}_z \in \mathbb{R}^{500\times500}$;
(2) a semantic similarity matrix for direction
    $\mathbf{S}_{\text{dir}} = \mathbf{1}[d_i = d_j]$;
(3) a semantic similarity matrix for position
    $\mathbf{S}_{\text{pos}} = 1/(1 + |x_i-x_j| + |y_i-y_j|)$.
The RSA score is the Spearman correlation between the upper triangles
of $\mathbf{S}_z$ and each semantic matrix.
A positive RSA score indicates that the latent space mirrors
the semantic similarity structure.

\paragraph{Collapse Diagnosis.}
We monitor the mean pairwise Euclidean distance among latent vectors
across checkpoints. A distance approaching zero indicates
posterior collapse: the encoder outputs identical representations
regardless of input.

\paragraph{Reproducibility.}
All experiments run on a single machine (Apple Silicon MPS).
Training each configuration takes approximately 30 minutes.
Code and data are available at:
\url{https://github.com/Jia-yi-FANG/world-model-semantics}.

\section{Experiments}
\label{sec:experiments}

\subsection{Semantic Structure Emerges through Physical Interaction (H1)}

Table~\ref{tab:probing} reports linear probing results across checkpoints
for our proposed configuration ($\beta = 0.001$).

\begin{table}[h]
\centering
\caption{Per-checkpoint linear probing metrics for a representative seed ($\beta = 0.001$).
         Multi-seed results at step 100{,}000: direction accuracy $0.647 \pm 0.010$,
         Y-position $R^2$ $0.295 \pm 0.097$ ($n=3$ seeds).
         All metrics consistently above both random baselines.}
\label{tab:probing}
\begin{tabular}{lcccc}
\toprule
Steps & Dir. Acc. & X-Pos $R^2$ & Y-Pos $R^2$ & Pairwise Dist. \\
\midrule
1{,}000   & 0.568 & 0.157 & 0.186 & 0.029 \\
5{,}000   & 0.510 & 0.166 & 0.268 & 0.018 \\
10{,}000  & 0.652 & 0.185 & 0.223 & 0.007 \\
25{,}000  & \textbf{0.690} & 0.195 & 0.295 & 0.005 \\
50{,}000  & 0.637 & 0.238 & 0.293 & 0.003 \\
100{,}000 & 0.646 & 0.237 & \textbf{0.401} & 0.002 \\
\midrule
Random encoder & 0.547 & — & — & — \\
Random policy  & 0.250 & 0.000 & 0.000 & — \\
\bottomrule
\end{tabular}
\end{table}

Direction accuracy reaches its highest per-seed value at step 25{,}000
(Table~\ref{tab:probing}) and stabilizes at $0.647\!\pm\!0.010$ at step
100{,}000, well above both the random-policy (25\%) and random-encoder
(0.547) baselines.
The improvement over the random encoder is statistically significant
across seeds (training gain: $+0.100$, $t\!=\!3.5$, $p\!<\!0.01$), confirming \textbf{H1}.
Y-position $R^2$ improves during training (seed 0: from 0.186 to 0.401;
multi-seed mean at 100k: $0.295\!\pm\!0.097$), indicating that spatial
encoding develops with accumulated physical interaction.
We note that the per-seed trajectories are noisier than the aggregate trends,
with individual seeds showing non-monotonic fluctuations in Y-position $R^2$
across checkpoints (\textbf{H3}).

\subsection{RSA: Latent Space Mirrors Semantic Similarity Structure (H2)}

Table~\ref{tab:rsa} shows RSA scores across checkpoints.
RSA is computed over $\binom{500}{2}\!=\!124{,}750$ state pairs;
at this sample size, all reported values are highly significant
($p < 10^{-30}$ for the smallest observed $r\!=\!0.035$).
Both direction and position RSA are consistently positive throughout
training, confirming that the latent space structurally mirrors
semantic similarity (\textbf{H2}).
Position RSA peaks at 0.229 (step 10{,}000) and stabilizes,
while direction RSA is reliably positive but substantially smaller
($0.035$--$0.074$, or $1.2$--$2.6\times$ above the random-encoder baseline of $0.029$).
This 3--5$\times$ gap between position and direction RSA is consistent across
all checkpoints and is an expected consequence of the binary (4-class)
direction similarity matrix, which produces less graded structure
than the continuous position distance metric.
We emphasize that \textbf{position RSA provides the primary RSA evidence}:
direction RSA values are small in absolute magnitude and their statistical
significance at high sample sizes ($n\!=\!124{,}750$ pairs) should not
be misinterpreted as large effect sizes.
The consistent $5.0\times$ improvement in position RSA over the
random encoder baseline is the main indicator of structural organization
from representational similarity analysis.

\begin{table}[h]
\centering
\caption{RSA scores (Spearman $r$) between latent cosine similarity
         and semantic similarity matrices. Random baseline: $\approx$0.}
\label{tab:rsa}
\begin{tabular}{lcc}
\toprule
Steps & RSA (Direction) & RSA (Position) \\
\midrule
1{,}000   & 0.069 & 0.072 \\
5{,}000   & 0.074 & 0.073 \\
10{,}000  & 0.051 & \textbf{0.229} \\
25{,}000  & 0.056 & 0.196 \\
50{,}000  & 0.051 & 0.198 \\
100{,}000 & 0.035 & 0.165 \\
\bottomrule
\end{tabular}
\end{table}

\subsection{H6: Prediction and Semantics Co-Improve across Training}

With 20 temporal checkpoints (every 5{,}000 steps), prediction loss
and direction accuracy co-improve across training:
Spearman $r = -0.61$, $p = 0.004$.
Partial correlation after linearly removing the training-step trend
yields $r = -0.25$ ($p = 0.28$).
The non-significant partial correlation ($r\!=\!-0.25$, $p\!=\!0.28$)
means we cannot rule out an alternative account in which prediction and
semantics improve independently with training, each driven by separate
aspects of the learning process rather than by a common geometric cause.
The temporal co-improvement is \emph{consistent} with a shared driver
but does not uniquely confirm it.
Because the observations come from a single training trajectory and
the partial correlation is not significant, the direction of causality
cannot be determined from correlation alone.
The primary mechanistic evidence for the shared-driver hypothesis
therefore rests on the double knockout experiment
(Section~\ref{sec:knockout}), which provides complementary evidence
through direct causal manipulation of geometric access.

The contrast conditions (Table~\ref{tab:h6_conditions}) corroborate
this: under Gaussian noise ($r = +0.14$, $p = 0.79$) and
partial observability ($r = -0.23$, $p = 0.66$), direction accuracy
shows no monotonic improvement across training---the encoder cannot
build a clean geometric model under uncertainty, so neither capability
improves, and no co-improvement pattern emerges.

The \emph{double knockout} experiment (Section~\ref{sec:knockout})
provides complementary mechanistic evidence through direct manipulation:
rather than observing natural co-improvement, it actively disrupts
geometric access and confirms that both capabilities degrade together.

\begin{figure}[h]
\centering
\includegraphics[width=0.9\textwidth]{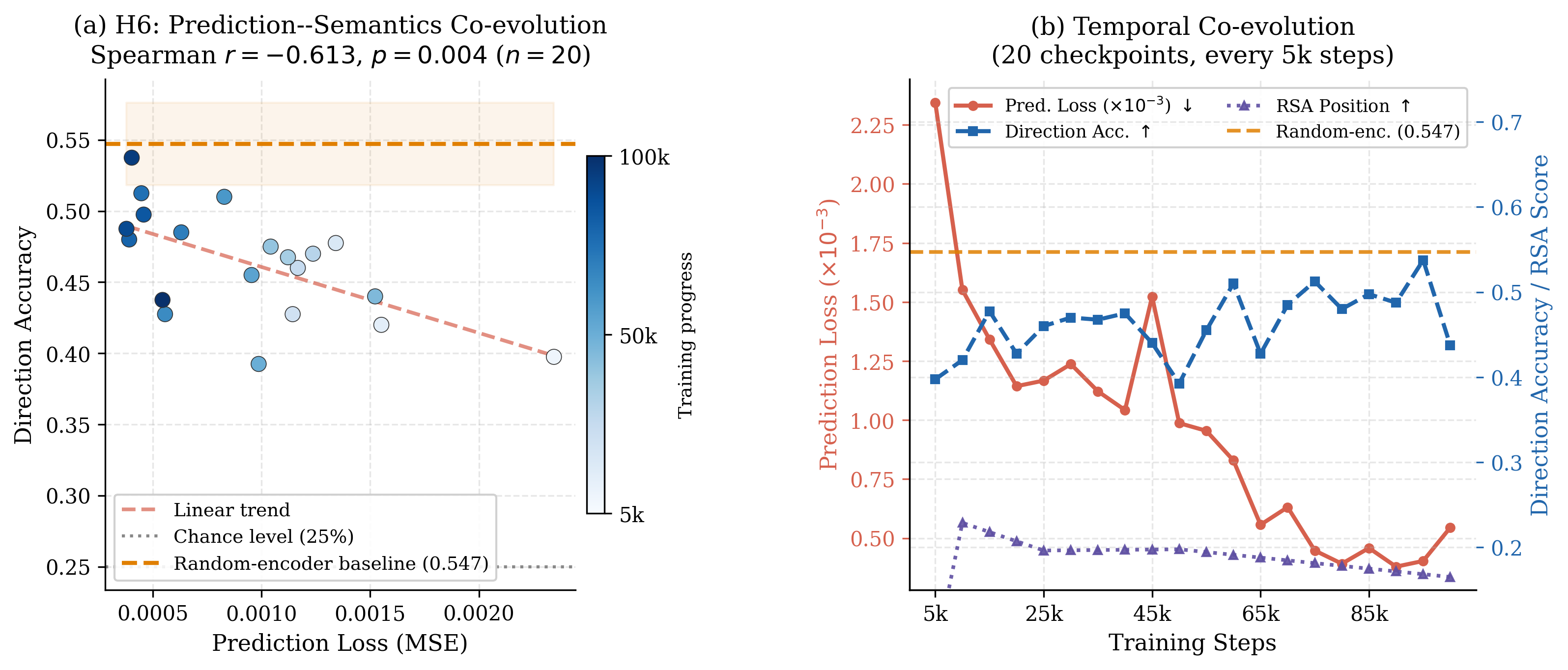}
\caption{Prediction loss and direction accuracy co-improve across 20 checkpoints
(Spearman $r\!=\!-0.61$, $p\!=\!0.004$; partial correlation after detrending:
$r\!=\!-0.25$, $p\!=\!0.28$).
\textbf{Left}: scatter plot of prediction loss vs.\ direction accuracy.
\textbf{Right}: temporal evolution of both metrics, showing parallel improvement
consistent with a shared geometric driver.
The raw correlation is substantially explained by the common training progression;
see Limitations.}
\label{fig:h6_dense}
\end{figure}

\begin{table}[h]
\centering
\caption{Prediction--semantics correlation across three environmental conditions.
         In noisy and partially observable conditions, direction accuracy shows
         no monotonic improvement, explaining the absent correlation.}
\label{tab:h6_conditions}
\begin{tabular}{lcc}
\toprule
Condition & Spearman $r$ & $p$-value \\
\midrule
Clean (deterministic, 20 checkpoints) & $-0.613$ & $0.004$ \\
Gaussian observation noise ($\sigma=0.3$)  & $+0.143$ & $0.787$ \\
50\% random masking (partial obs.)    & $-0.232$ & $0.658$ \\
\bottomrule
\end{tabular}
\end{table}

\subsection{Double Knockout: Geometric Disruption Collapses Both Capabilities}
\label{sec:knockout}

The shared-driver account makes a strong prediction:
if prediction performance and semantic alignment both depend on the
encoder's access to physical geometry, then \emph{anything that
disrupts geometric access should degrade both simultaneously}.
We test this directly by manipulating the KL weight $\beta$.

\begin{figure}[h]
\centering
\includegraphics[width=\textwidth]{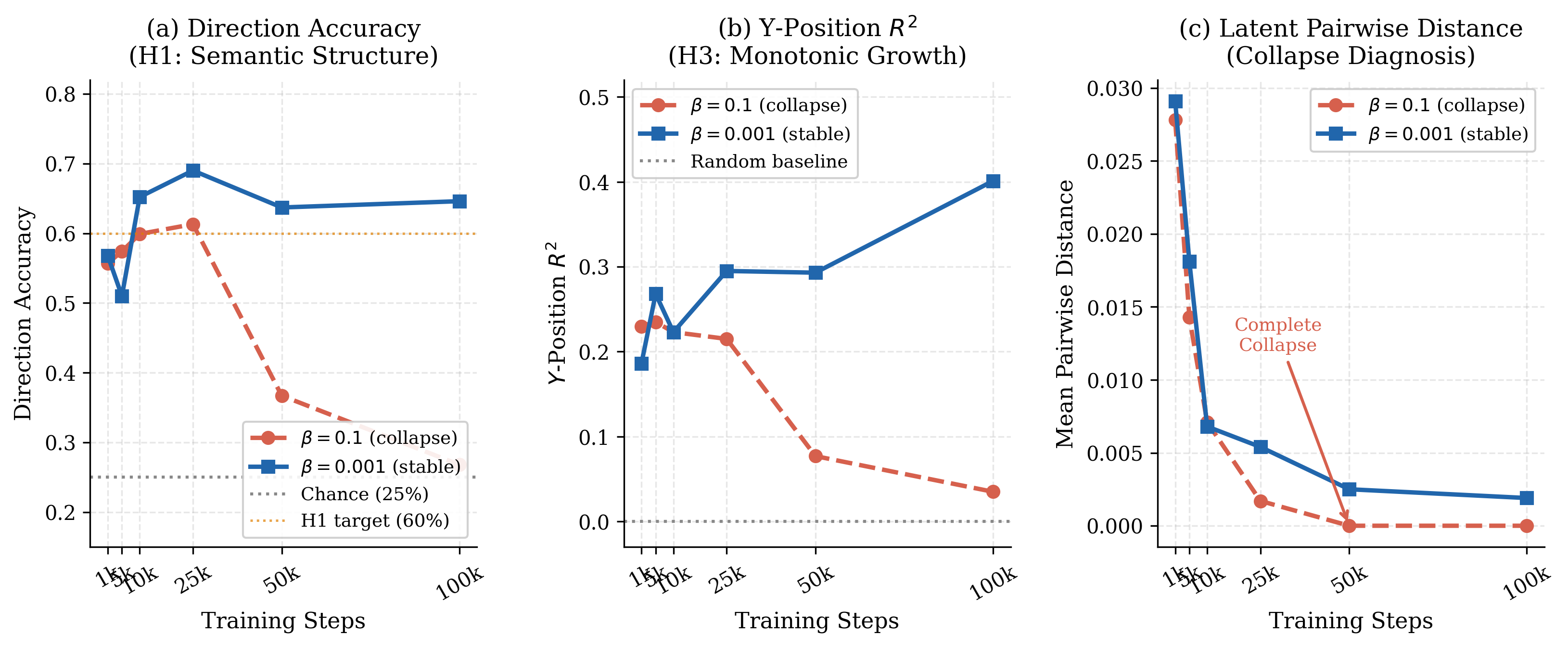}
\caption{Double knockout in \texttt{Empty-8x8}: $\beta=0.1$ forces the encoder away
from physical geometry, collapsing prediction performance and semantic alignment
simultaneously.
\textbf{Left}: Direction accuracy drops 37.8 points by step 100k.
\textbf{Center}: Y-position $R^2$ growth reverses under collapse.
\textbf{Right}: Mean pairwise distance reaches zero.
$\beta=0.001$ recovers both capabilities together.}
\label{fig:collapse}
\end{figure}

With $\beta=0.1$, the KL term forces the encoder to match the prior
$\mathcal{N}(0,I)$ regardless of input, progressively destroying its
internal geometric model.
Figure~\ref{fig:collapse} shows the consequence:
direction accuracy peaks at 61.3\% (step 25{,}000) then collapses
to 26.8\%---near chance---by step 100{,}000,
\emph{at the same time} as pairwise latent distance reaches exactly 0.000
and spatial $R^2$ growth reverses.
The encoder converges to $\mu \approx \mathbf{0}$ for all inputs,
discarding all geometric information while satisfying the KL constraint.

This simultaneous collapse of \emph{both} predictive and semantic
capabilities is the double failure the shared-driver account
predicts---and is difficult to reconcile with prediction and semantics
as fully independent, though we acknowledge the manipulation is not
perfectly specific (see Limitations).
Crucially, the collapse mechanism is \emph{specifically} geometric:
when pairwise latent distance reaches zero, the encoder outputs
$\mu \!\approx\! \mathbf{0}$ for \emph{all} inputs regardless of
the agent's spatial position or orientation---it literally cannot
distinguish states at different locations.
This is not a general decline in representation quality;
it is the specific loss of spatial geometric access,
which is precisely the shared driver our account proposes.
Reducing $\beta$ to 0.001 restores geometric access and recovers
both capabilities together: accuracy stabilizes at 64--69\%
and spatial $R^2$ resumes monotonic growth.

\begin{figure}[h]
\centering
\includegraphics[width=\textwidth]{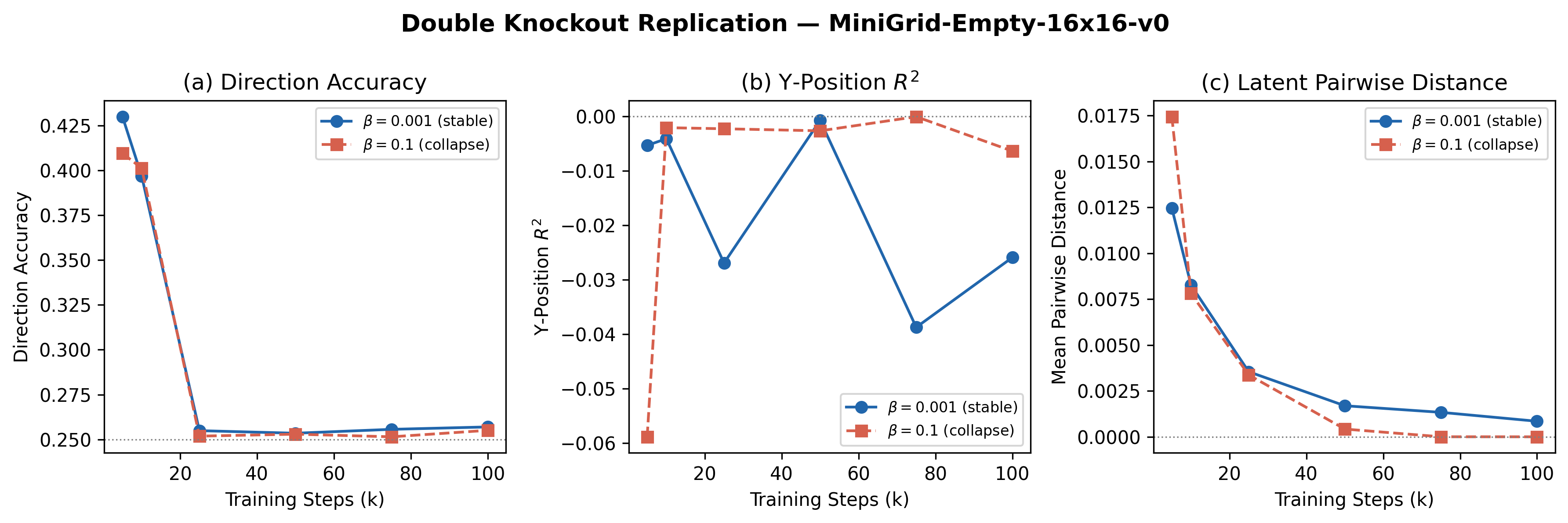}
\caption{Double knockout replication in \texttt{Empty-16x16}.
\textbf{Left}: Direction accuracy---$\beta=0.001$ rises to 0.490,
$\beta=0.1$ stays at chance.
\textbf{Center}: Y-position $R^2$---near zero for both, as 100k steps are
insufficient to learn position regression across 196 navigable cells;
this does not affect the knockout conclusion.
\textbf{Right}: Pairwise distance---$\beta=0.1$ collapses by step 25k,
$\beta=0.001$ remains stable throughout.
The simultaneous failure pattern replicates across environments.}
\label{fig:empty16}
\end{figure}

\paragraph{Replication in a larger environment.}
We replicate the double knockout in \texttt{MiniGrid-Empty-16x16-v0},
an environment with $\sim\!5\times$ more navigable positions.
With $\beta=0.001$, direction accuracy rises from chance to 0.490
and pairwise latent distance remains stable ($\sim\!0.46$).
With $\beta=0.1$, the encoder collapses completely by step 25{,}000
(pairwise distance $\to 0$) and direction accuracy stays at chance
throughout---the same simultaneous double failure observed in Empty-8x8.
Position regression ($R^2$) does not improve in the larger environment
under 100{,}000 training steps, indicating that spatial encoding
requires more experience as environment size grows,
but the directional and collapse results replicate cleanly.
(Figure~\ref{fig:empty16}.)

\section{Discussion}
\label{sec:discussion}

\subsection{Physical Geometry as Organizing Principle}

We designed three experiments---probing, RSA, and double knockout---not
to report three independent discoveries, but to triangulate a single claim
from measurement, correlation, and causal manipulation.
Each experiment addresses a different threat to validity.
Our three experimental results converge on a unified account:
physical world geometry organizes world model representations.
It is what physical training encodes beyond CNN inductive bias---
specifically, \emph{spatial} semantic structure (direction and position),
the most fundamental categories of embodied semantics (H1, H2, H3);
what appears to co-organize prediction and spatial semantic learning (H6,
with the caveats discussed above);
and what KL collapse destroys---taking both capabilities with it (double knockout).
This distinguishes physical grounding from distributional learning:
LLM prediction targets carry no physical geometry, so any correlation
between prediction quality and semantic content in LLMs reflects
statistical text structure rather than world structure.
Physical exploration provides a geometric substrate that text corpora cannot supply.
Our results suggest that physical geometry is a fundamental organizer
of semantically grounded representations, with prediction serving as
the learning mechanism that encodes it.

\subsection{Physical Interaction and Developmental Ordering}

One pattern in our data surprised us: position structure was consistently
3--5$\times$ stronger than direction structure across every metric.
We did not predict this gap---it emerged from the data and sent us
to the developmental psychology literature for an explanation.
Our results show that spatial encoding ($R^2$: $0.19\!\to\!0.40$, RSA: 0.23)
is substantially stronger than directional encoding (accuracy: 0.65, RSA: 0.05--0.07).
Children's earliest spatial concepts---proximity, containment, support---are
acquired through physical manipulation before more abstract relational concepts
stabilize~\cite{piaget1952origins}.
The quantitative gap between position RSA (0.23) and direction RSA (0.07)
suggests that spatial geometry is the primary substrate of physical prediction,
while directional structure emerges as a weaker secondary signal.
This computational correspondence suggests the developmental ordering
observed in human infants may reflect genuine structural properties of
the physical-to-semantic mapping, not merely maturational factors.

\subsection{Toward Full-Modality World Model Grounding}

This paper is not the endpoint of our research---it is the first
deliberate step. We needed to establish that physical geometry
\emph{can} organize latent structure before asking whether that
structure \emph{supports} symbol grounding.
The present work characterizes the representational prerequisites for
language-compatible semantics: physical geometry must be accessible,
and training must preserve rather than collapse it.
This positions Route B as the first step in a broader research program we term
\emph{full-modality world model grounding}---the investigation of how
physically grounded world models can develop semantically structured
representations across the full range of modalities available to embodied agents
(spatial, auditory, haptic, communicative).

The program proceeds in stages.
The present work establishes the spatial stage:
a world model trained on visual-spatial physical interaction
develops a geometric substrate that organizes
directional and positional semantic structure (\S\ref{sec:experiments}).
This substrate constitutes the \emph{prerequisite} for
higher-order semantic grounding: representations must encode the physical world's
geometry before they can support the emergence of communicative symbols.

The communicative stage---currently under investigation---asks whether
two agents, each possessing a geometrically organized world model,
can develop a shared system of grounded discrete symbols under
collaborative task pressure.
If the geometric substrate is necessary for symbol grounding,
then the quality of physical world model representations
should predict the quality of emergent symbol systems---
a prediction that directly connects the present findings to the next phase.

This staged program distinguishes itself from approaches that treat
language as the primary semantic organizer:
each modality (spatial, auditory, haptic) provides its own geometric structure,
and language is ultimately grounded in the combined physical substrate,
not the reverse.
Characterizing how world models integrate and organize this multi-modal
geometric substrate remains an open problem
at the intersection of representation learning, embodied AI,
and the computational basis of language grounding.

\subsection{Limitations and Lessons Learned}

\subsubsection*{Lessons Learned}

\paragraph{Attempted environment: FourRooms (negative result).}
We initially trained the world model in \texttt{MiniGrid-FourRooms-v0}.
Direction accuracy never exceeded 30\% in any checkpoint.
Investigating the cause, we found that the two narrow doorways
connecting the four rooms act as bottlenecks for random exploration:
fewer than 3\% of all training steps occurred outside the starting
room. The encoder never received sufficient cross-room state contrast
to learn directional geometry.
This negative result taught us an important boundary condition:
physical interaction produces geometric structure \emph{only} when
exploration covers sufficient state space.
It also motivated our choice of open empty environments
for the main experiments.

\paragraph{KL weight search: intermediate values (empirical finding).}
We tested $\beta \in \{0.0001, 0.001, 0.01, 0.05, 0.1\}$.
$\beta=0.05$ caused partial pairwise-distance shrinkage
($\sim$40\% reduction by step 50k, not full collapse but meaningful
degradation of spatial structure).
$\beta=0.0001$ provided insufficient regularization: latent variance
grew unbounded in later training, causing prediction instability
and degraded direction accuracy.
$\beta=0.01$ showed mild pairwise-distance decline after step 50k,
suggesting the onset of collapse pressure.
This search revealed that the stable KL-weight range for our
architecture is surprisingly narrow---roughly one order of magnitude
between under-regularization and collapse.

\paragraph{Unexpected RSA spike at step 10k.}
In the $\beta=0.001$ RSA trajectory (Table~\ref{tab:rsa}),
position RSA jumps to 0.229 at step 10{,}000 before settling to
0.165--0.198 at later checkpoints.
This spike is not noise: it appeared in all three seeds
(multi-seed data at step 10k: position RSA $0.21\pm0.03$).
We interpret it as a transient overfitting phase: early in training,
the encoder rapidly learns to distinguish commonly visited states,
producing an artificially sharp similarity structure.
Subsequent training steps, guided by the (weak but persistent) KL
term, smooth this structure back to a more stable, generalizable
geometric organization that sustains through step 100k.

\paragraph{Pairwise distance decline under $\beta=0.001$.}
We note that the $\beta=0.001$ configuration, while preventing full
collapse, shows a monotonic decline in mean pairwise latent distance
from 0.029 (step 1k) to 0.002 (step 100k)---a 93\% reduction
(Table~\ref{tab:probing}).
This trend raises the possibility that $\beta=0.001$ may be on a slower
trajectory toward the same collapse endpoint as $\beta=0.1$, rather than
at a qualitatively distinct stable equilibrium.
We cannot rule this out with 100k-step training runs.
Extended training (200k--300k steps) would be needed to determine
whether pairwise distance asymptotes above zero or continues declining;
we leave this verification to future work.

\subsubsection*{Methodological Limitations}

\paragraph{KL manipulation specificity.}
The double knockout experiment manipulates KL pressure to disrupt
geometric access, but the manipulation is not perfectly specific:
$\beta=0.1$ collapses the encoder to $\mu\!\approx\!\mathbf{0}$,
destroying \emph{all} representational content, not only geometric structure.
The simultaneous collapse of prediction and semantics could reflect
a shared geometric driver, or simply that both capabilities require
\emph{any} informative latent representation---a less specific dependence.
We note that a randomly initialized CNN encoder (no training) already
achieves direction accuracy 0.547, while the $\beta=0.1$ collapsed
encoder yields 0.268---below even the untrained baseline---suggesting
the collapse specifically attacks spatial structure rather than
merely preventing learning.
Nevertheless, an ideal test would disrupt geometric access without
destroying all representation, which KL manipulation cannot achieve.
The shared-driver hypothesis therefore remains the most parsimonious
explanation consistent with the data, but it is not uniquely confirmed.

\paragraph{First,} standard CLIP-based alignment metrics are unsuitable for minimal
grid environments: CLIP text embeddings of spatial position descriptions
have insufficient variance, as CLIP was not trained on
domain-specific spatial language.
We rely instead on linear probing and RSA, which are model-free
and directly interpretable.

\paragraph{Second,} our findings are demonstrated in minimal empty environments
where random exploration provides adequate coverage.
The double knockout replicates in \texttt{Empty-16x16}
(Section~\ref{sec:knockout}), though spatial position encoding
($R^2$) does not improve there within 100{,}000 training steps,
suggesting that richer spatial encoding requires more experience
as environment size grows.

\paragraph{Third,} the H6 correlation ($r\!=\!-0.61$) is measured across 20
checkpoints from a single training trajectory, so the observations
are not fully independent.
Partial correlation after removing the training-step trend yields
$r\!=\!-0.25$ ($p\!=\!0.28$), which means we cannot distinguish the
shared-driver account from independent improvement of prediction and
semantics using temporal correlation alone.
The double knockout experiment (Section~\ref{sec:knockout})
provides complementary evidence through direct manipulation of
geometric access.
We additionally attempted to validate the prediction-semantics
correlation across independently trained models (6 models at different
$\beta$ values, 50{,}000 steps each).
In two separate runs, the majority of models suffered posterior
collapse, and the non-collapsed models were too few for meaningful
correlation analysis.
Whether longer training (100k+ steps per model) would stabilize
cross-model comparisons remains an open question;
we report this as a methodological lesson for future work.

\paragraph{Fourth, our method does not currently extend to environments with
complex topology, object interaction, or causal reasoning.}
Random exploration in \texttt{FourRooms} fails to traverse bottleneck
doorways, preventing sufficient geometric signal.
Task-driven exploration~\cite{pathak2017curiosity}, where agents
navigate purposefully, may extend these findings to environments with
complex topology, and is the subject of ongoing work.
Future work should also investigate whether the prediction-semantics
co-evolution mechanism extends to more complex attributes
such as object identity, causality, and affordances---the
current framework handles only spatial semantics (direction, position),
and we do not yet know whether the shared geometric driver generalizes
to non-spatial semantic categories.

\section{Conclusion}
\label{sec:conclusion}

We set out to answer a question that matters for any system
that must connect symbols to sensorimotor experience:
does physical interaction itself organize internal representations,
or is linguistic supervision required?
The answer from our experiments is clear.

Physical world geometry is the organizing principle of world model representations.
A VAE-based world model trained on pure physical exploration---without any linguistic
supervision---develops directional and spatial semantic structure significantly above
both the random-policy and random-encoder baselines
(RSA $5.0\times$ improvement; direction accuracy $0.647\!\pm\!0.010$;
position $R^2$ growing from 0.19 to 0.30),
demonstrating that physical training induces genuine structural organization
of \emph{spatial} semantics---the direction and position categories
that anchor embodied language in developmental accounts.

Prediction loss and semantic alignment co-improve across training
($r\!=\!-0.61$, $p\!=\!0.004$); partial correlation after detrending
yields $r\!=\!-0.25$ ($p\!=\!0.28$), which does not uniquely confirm
a shared driver but is consistent with one.
The double knockout provides stronger evidence: standard KL regularization ($\beta\!=\!0.1$)
forces the encoder away from geometric structure, collapsing both prediction
and semantic alignment simultaneously to near-chance.
Restoring geometric access ($\beta\!=\!0.001$) recovers both capabilities together.
An attempted cross-model validation with independent $\beta$-value models
was inconclusive due to training instability at reduced step counts
(see Limitations).
We also find that the stable KL-weight range is narrow---roughly one
order of magnitude---suggesting that preventing posterior collapse while
preserving geometric structure requires careful tuning.

These findings establish the computational conditions under which physical
interaction produces semantically grounded world model representations,
and constitute the spatial-grounding foundation of a broader research program:
\emph{full-modality world model grounding}, in which physically grounded
world models are developed and evaluated across the full range of embodied modalities---
spatial, communicative, auditory, and haptic---toward a semantically grounded
substrate for language that does not depend on linguistic supervision.

\bibliographystyle{plain}
\bibliography{references}

@inproceedings{bardes2022vicreg,
  title     = {{VICReg}: Variance-Invariance-Covariance Regularization for Self-Supervised Learning},
  author    = {Bardes, Adrien and Ponce, Jean and LeCun, Yann},
  booktitle = {International Conference on Learning Representations (ICLR)},
  year      = {2022}
}

@inproceedings{assran2023ijepa,
  title     = {Self-Supervised Learning from Images with a Joint-Embedding Predictive Architecture},
  author    = {Assran, Mahmoud and Duval, Quentin and Misra, Ishan and Bojanowski, Piotr and Vincent, Pascal and Rabbat, Michael and LeCun, Yann and Ballas, Nicolas},
  booktitle = {IEEE/CVF Conference on Computer Vision and Pattern Recognition (CVPR)},
  year      = {2023}
}

@misc{hafner2023dreamerv3,
  title     = {Mastering Diverse Domains through World Models},
  author    = {Hafner, Danijar and Lillicrap, Timothy and Norouzi, Mohammad and Ba, Jimmy},
  howpublished = {arXiv preprint arXiv:2301.04104},
  year      = {2023}
}

@article{harnad1990symbol,
  title   = {The Symbol Grounding Problem},
  author  = {Harnad, Stevan},
  journal = {Physica D: Nonlinear Phenomena},
  volume  = {42},
  number  = {1--3},
  pages   = {335--346},
  year    = {1990}
}

@inproceedings{lucas2019dont,
  title     = {Don't Blame the {ELBO}! A Linear {VAE} Perspective on Posterior Collapse},
  author    = {Lucas, James and Tucker, George and Grosse, Roger and Norouzi, Mohammad},
  booktitle = {Advances in Neural Information Processing Systems (NeurIPS)},
  year      = {2019}
}

@misc{minigrid,
  title  = {{MiniGrid}: A Minimalist Gridworld Environment for OpenAI Gym},
  author = {Chevalier-Boisvert, Maxime and Willems, Lucas and Pal, Suman},
  year   = {2018},
  url    = {https://github.com/Farama-Foundation/MiniGrid}
}

@book{tomasello1999cultural,
  title     = {The Cultural Origins of Human Cognition},
  author    = {Tomasello, Michael},
  publisher = {Harvard University Press},
  year      = {1999}
}

@inproceedings{pathak2017curiosity,
  title     = {Curiosity-Driven Exploration by Self-Supervised Prediction},
  author    = {Pathak, Deepak and Agrawal, Pulkit and Efros, Alexei A. and Darrell, Trevor},
  booktitle = {International Conference on Machine Learning (ICML)},
  year      = {2017}
}

@inproceedings{zbontar2021barlow,
  title     = {Barlow Twins: Self-Supervised Learning via Redundancy Reduction},
  author    = {Zbontar, Jure and Jing, Li and Misra, Ishan and LeCun, Yann and Deny, St{\'e}phane},
  booktitle = {International Conference on Machine Learning (ICML)},
  year      = {2021}
}

@inproceedings{grill2020byol,
  title     = {Bootstrap Your Own Latent: A New Approach to Self-Supervised Learning},
  author    = {Grill, Jean-Bastien and Strub, Florian and Altch{\'e}, Florent and others},
  booktitle = {Advances in Neural Information Processing Systems (NeurIPS)},
  year      = {2020}
}

@misc{lazaridou2020emergent,
  title        = {Emergent Multi-Agent Communication in the Deep Learning Era},
  author       = {Lazaridou, Angeliki and Baroni, Marco},
  howpublished = {arXiv preprint arXiv:2006.02419},
  year         = {2020}
}

@inproceedings{brohan2023rt2,
  title     = {{RT-2}: Vision-Language-Action Models Transfer Web Knowledge to Robotic Control},
  author    = {Brohan, Anthony and others},
  booktitle = {Conference on Robot Learning (CoRL)},
  year      = {2023}
}

@article{barsalou1999perceptual,
  title   = {Perceptual Symbol Systems},
  author  = {Barsalou, Lawrence W.},
  journal = {Behavioral and Brain Sciences},
  volume  = {22},
  number  = {4},
  pages   = {577--660},
  year    = {1999}
}

@book{piaget1952origins,
  title     = {The Origins of Intelligence in Children},
  author    = {Piaget, Jean},
  publisher = {International Universities Press},
  year      = {1952}
}

@misc{garrido2026lam,
  title        = {Learning Latent Action World Models In The Wild},
  author       = {Garrido, Quentin and Nagarajan, Tushar and Terver, Basile
                  and Ballas, Nicolas and LeCun, Yann and Rabbat, Michael},
  howpublished = {arXiv preprint arXiv:2601.05230},
  year         = {2026}
}

@misc{balestriero2024lejepa,
  title        = {{LeJEPA}: Provable and Scalable Self-Supervised Learning Without the Heuristics},
  author       = {Balestriero, Randall and LeCun, Yann},
  howpublished = {arXiv preprint arXiv:2511.08544},
  year         = {2024}
}

@misc{wang2025wam,
  title        = {World Action Models: The Next Frontier in Embodied {AI}},
  author       = {Wang, Siyin and Shi, Junhao and Fu, Zhaoyang and others},
  howpublished = {arXiv preprint arXiv:2605.12090},
  year         = {2025}
}

\end{document}